\documentclass[lettersize,journal]{IEEEtran}
\usepackage{amsmath,amsfonts}
\usepackage{algorithm}
\usepackage{algpseudocode}
\usepackage{array}
\usepackage[caption=false,font=normalsize,labelfont=sf,textfont=sf]{subfig}
\usepackage{textcomp}
\usepackage{stfloats}
\usepackage{url}
\usepackage{verbatim}
\usepackage{graphicx}
\usepackage{cite}
\usepackage{subfig}

\usepackage{multirow}
\usepackage{booktabs}       
\usepackage[pdfstartview=XYZ,
bookmarks=true,
colorlinks=true,
linkcolor=blue,
urlcolor=blue,
citecolor=blue,
pdftex,
bookmarks=true,
linktocpage=true, 
hyperindex=true
]{hyperref}

\usepackage{orcidlink}

\begin{document}

\title{Efficacy of Neural Prediction-Based Zero-Shot NAS}

\author{Minh Le\,\orcidlink{0000-0002-1570-5754}     
    \and
    Nhan Nguyen\,\orcidlink{0009-0004-4654-6778}
    \and
    Ngoc Hoang Luong\,\orcidlink{0000-0002-6768-1950}}


\maketitle

\begin{abstract}
In prediction-based Neural Architecture Search (NAS), performance indicators derived from graph convolutional networks have shown remarkable success. These indicators, achieved by representing feed-forward structures as component graphs through one-hot encoding, face a limitation: their inability to evaluate architecture performance across varying search spaces. In contrast, handcrafted performance indicators (zero-shot NAS), which use the same architecture with random initialization, can generalize across multiple search spaces. Addressing this limitation, we propose a novel approach for zero-shot NAS using deep learning. Our method employs Fourier sum of sines encoding for convolutional kernels, enabling the construction of a computational feed-forward graph with a structure similar to the architecture under evaluation. These encodings are learnable and offer a comprehensive view of the architecture's topological information. An accompanying multi-layer perceptron (MLP) then ranks these architectures based on their encodings. Experimental results show that our approach surpasses previous methods using graph convolutional networks in terms of correlation on the NAS-Bench-201 dataset and exhibits a higher convergence rate. Moreover, our extracted feature representation trained on each NAS benchmark is transferable to other NAS benchmarks, showing promising generalizability across multiple search spaces. The code is available at: 

\url{https://github.com/minh1409/DFT-NPZS-NAS}
\end{abstract}


%
\IEEEpeerreviewmaketitle

\section{Introduction}

Neural Architecture Search (NAS) aims to automate the discovery of top-performing neural network architectures via substituting or enhancing the expert-driven design process with an algorithmic exploration mechanism \cite{nassucess1, nassucess2, nassucess3, nassucess4}. Since traditional NAS procedures require models to be fully trained, they cost a lot of computational resources. To reduce the computational burden of NAS procedures, two promising paradigms have emerged: Prediction-based NAS \cite{learningcurve, sppedup, nasbot, BRP-NAS, TNASP} and Zero-Shot NAS \cite{naswot, TE-NAS, LightweightNAS, Zen-NAS}.

Prediction-based NAS is created to evaluate the performance of fewer neural networks by efficiently interpret the data. The research is currently dominated by neural models \cite{BRP-NAS, Gate}. They are able to evaluate the performance of architectures within a search space, using only a small sample from that search space. It implies that neural networks can generalize extremely well within the search space consisting of the sample these neural networks learned from. However, the learned representation from prediction-based NAS cannot be reused to search in another search space due to some inherent differences between these spaces, making a prediction-based NAS process costly in many applications. This limitation of prediction-based NAS is largely based on different NAS search spaces using different types of layers while most prediction-based NAS relies on one-hot encoding to represent these layers (all current graph-learning-based, for example: \cite{BRP-NAS, TNASP}).

Zero-Shot NAS is created to consume less resource per evaluated architectures by being able to measure the performance of architectures without training them. To conduct architecture performance measurement, Zero-Shot NAS procedures utilize a performance estimator or an ensemble of these. These performance estimators are crafted through expert knowledge based on empirical results. Simple forms of performance can be the depth or the width of neural network architectures, while modern performance estimators are based on insights on trainability and expressivity. These estimators are quick to compute, usually cost around one forward pass. If they can accurately predict the performance of a neural network, they can speed up the design process. In the meantime, while research is being conducted to improve zero-shot NAS, it does not currently have comparable accuracy to prediction-based NAS~\cite{BRP-NAS, naswot}.

Moreover, many of them rely on well-defined features like capacity, expressivity, and SGD-trainability \cite{naswot, Zen-NAS, TE-NAS}. These features, easy to understand at first glance, have a strong positive link with how well neural networks work. But to build a theory around neural network performance or design the architecture by hand, we need deep expertise. Often, handcrafted architectures are designed to boost certain metrics, like having bigger kernel sizes, wider models, or more layers. Therefore, the process of making handcraft neural network architecture performance predictors is the encapsulation of expert knowledge. However, relying on expert knowledge to design neural networks does not align with automated discovery of neural network architectures. This is a natural question: ``\textbf{Can neural networks interpret neural network architecture performance?}'' Using machine learning models to understand this falls under Prediction-based NAS, and there is a problem. The common method of using one-hot vectors to represent neural operators does not work well with the variety of neural operators. Therefore, we need a different encoding method for neural operators. We notice that each neural network gives a unique set of outputs carrying key structural details about the network, even if it looks like raw data that's tough to decipher at first. Given this, deep learning shows promise in helping us understand this data, especially since it's known for processing complex information.

\begin{figure}
        \centering
        \includegraphics[scale=0.29]{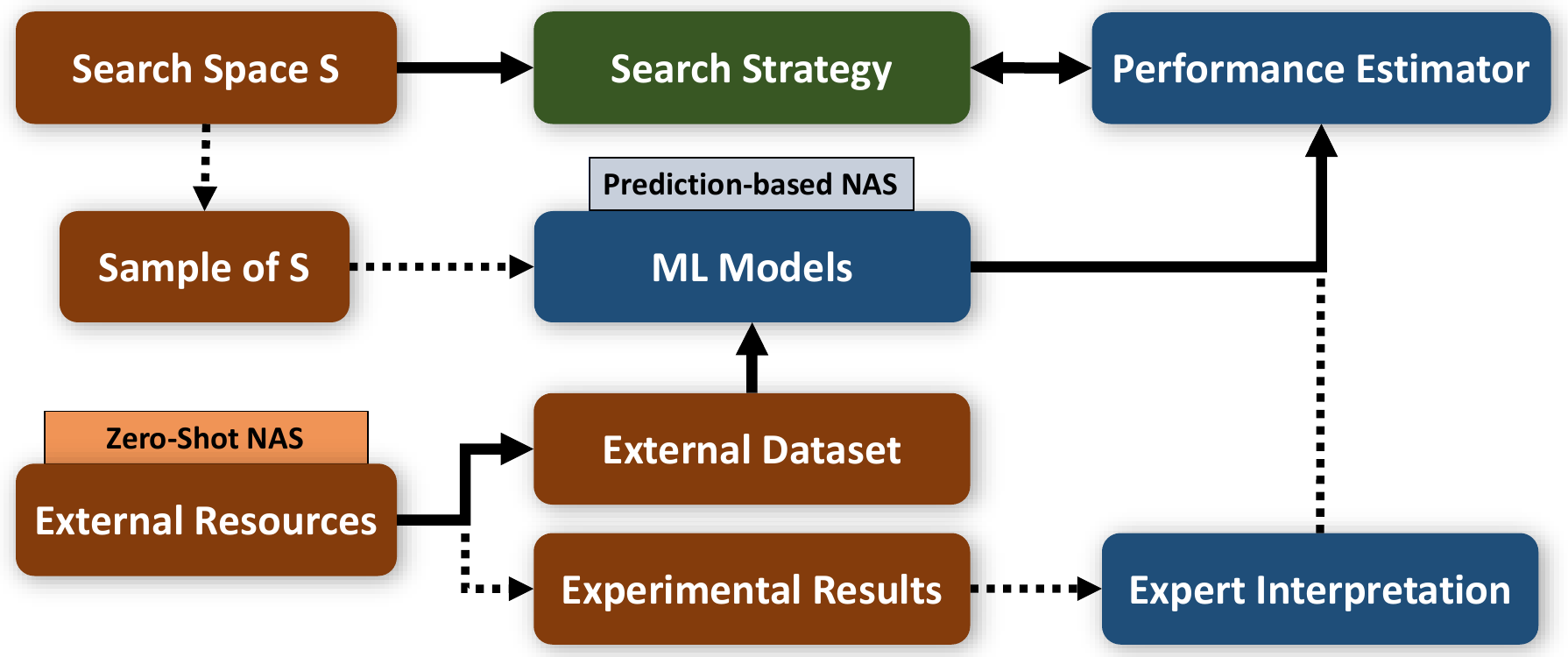}
        \caption{NAS Procedures with Different Performance Estimators}
        \label{fig:method}
    \end{figure}

In this study, we aim to show that neural prediction-based NAS can make good zero-shot NAS. Our procedure can be described in Figure \ref{fig:method}, where solid arrows and the dashed arrows describe the novel Prediction-based Zero-Shot NAS procedure and prior alternatives, respectively. We mitigate the limitation of prediction-based NAS paradigm we raised above by creating neural-based architectural performance predictors appropriate to a wide range of search space. By being able to use these neural-based architectural performance predictors, they can effectively function as performance predictors in Zero-Shot NAS paradigm. First, we propose a \emph{representation mechanism} to substitute neural network operators with components that resemble the operators being substituted. This design methodology can well-adapt future neural operations, for example, convolution represents convolution. Our proposed Fourier-Transform-based representation for convolution layers provides a procedure to encode all kinds of convolutional layers. It proved itself capable to capture topological information of the convolutional layer by achieving a higher score-accuracy correlation in prediction-based NAS for NAS-Bench-201 \cite{nb201}, many zero-shot NAS score-accuracy correlation test settings \cite{nds, nb201, nb101, macro}, and usefulness in conducting end-to-end NAS in a huge NAS search space \cite{Zen-NAS}. Second, we propose a new efficient optimization scheme that works for training on single and multiple architectural search spaces. The new optimization scheme improves the convergence rate of neural performance predictors. It allows them to be trained on large-scale datasets or to be trained within a small computational budget. Finally, we propose a way to ensemble multiple proxies, either to provide more reliable indicators or to mitigate visible weaknesses of certain performance predictors.

This paper is organized as follows. Section \ref{sec:related_work} presents existing work related to this article. Section \ref{sec:method} presents the mechanisms, optimization method, and search algorithm of NAS process. Section \ref{sec:experiment} shows the results of the proposed method in both prediction-based NAS and zero-shot NAS test settings in diverse benchmark and real search spaces. Section \ref{sec:discussion} provides certain interpretation of the results and limitations of the study. Finally, section \ref{sec:conclusion} presents the conclusions and future directions toward zero-shot NAS designed by machines.

\section{Related Work} \label{sec:related_work}

A typical NAS procedure consists of three components: search space, search strategy, and performance estimation strategy. The search strategy instructs the performance estimation strategy to evaluate an architecture from the search space. Then, it receives the estimation results to determine the next steps. Initial NAS research focused on the search strategy, primarily on designing heuristics tailored for NAS \cite{Genetic, RegularizedEvo} and deep learning-based methods \cite{rl1}. Efficient search space designs are also a popular NAS research topic \cite{nds, Mobilenetv3} because it prevents NAS procedure to evaluate obviously bad architectures existing in the search space. In this section, we provide reviews and summaries of works relevant to Prediction-based NAS and Zero-Shot NAS, which are directly related to our work.

\textbf {Zero-Shot NAS} are created to mitigate the computational burden of NAS. Traditional NAS procedures take prohibitively large computational resources, such as the one proposed by Zoph and Le \cite{rl1} costing 1800 GPU-days on a Tesla P100 GPU. Reliable performance estimators for neural network architectures actually existed before the emergence of NAS. They are the number of parameters and the depth of deep neural architectures. The belief in the superiority of big deep models over small shallow models has been the very first motivation to develop early deep learning architectures. It gave rise to models like LeNet-5 \cite{lenet}, AlexNet \cite{alexnet}, VGGNet \cite{vgg}. Expert knowledge about the architecture's expressivity, capacity, and SGD trainability has led to the creation of more sophisticated architecture performance predictors. Mellor et al. \cite{naswot} showed that NAS could be performed without actually optimizing the architecture. To estimate architecture performance at initialization, they measure the differences between binary codes of two different inputs. Their method uses the score as a performance evaluator for evolutionary algorithms. They achieved a moderate-to-strong correlation on NDS search spaces, on both CIFAR and ImageNet datasets. Chen et al. \cite{TE-NAS} rank architectures based on trainability and expressivity of architectures. Trainability of models refers to the ease of using gradient descent to optimize neural networks. They used condition number of neural tangent kernel (NTK) as an indicator of neural network trainability from the work of Xiao et al. \cite{Disentangling}. Expressivity of models refers to the complexity of models. To measure the complexity of models, they have used the number of linear regions, adapted from the work of Xiong et al. \cite{linearregion}. Their contribution lies in being the early Zero-Shot NAS, how to ensemble the two indicators, and how to search using their indicator. They shown that condition number of NTK and number of linear regions favors different type of operations: trainability favors skip connection and expressivity favors learnable operations on NAS-Bench-201 \cite{nb201}. As an effort to prevent one score completely dominate the other, they used the relative ranking instead of a linear combination because the search algorithm does not know the scale of two scores in the new search space. Their search algorithm is prune-based, called prune-by-importance. They have found SoTA on DARTS search space \cite{DARTS} as well as having an infinitesimal search cost. Abdelfattah et al. \cite{LightweightNAS} use metrics introduced by network pruning works to craft zero-shot performance estimators. The saliency-based metrics used in the paper are snip \cite{snip}, grasp \cite{grasp}, fisher \cite{fisher}, and synflow \cite{synflow}. They use the sum of saliency scores of all parameters as the performance predictor. As a consequence, the behavior of their proxies is similar to the number of parameters, explored in \cite{deeperlookzerocost}. They have run extensive experiments on and their method outperforms many prior performance estimators NAS-Benchmarks, namely NAS-Bench-201 \cite{nb201}, NAS-Bench-101 \cite{nb101}, and NAS-Bench-ASR \cite{nbasr}. Lin et al. \cite{Zen-NAS} introduce a scale-insensitive version (Zen-Score) of their $\Phi$-score as the expressivity of architecture. $\Phi$-score is calculated via expected gaussian complexity at each linear region. To make Zen-Score scale-insensitive, they re-scale $\Phi$-score by the product of the variance statistics of all the batch normalization (BN) layers in the network. They show an elegant method that maximizes architectural complexity and can discover high-performance networks in vanilla network space. Their method, however, \underline{cannot be used in irregular design spaces}, such as the one covered in NDS design spaces \cite{nds}. It discovered high-performance architectures in MobileNetV3 \cite{Mobilenetv3} search space and is the first Zero-Shot NAS to outperform prediction-based NAS on ImageNet.

\textbf{Prediction-based NAS} describes a class of methods leveraging architecture performance information obtained through training of architectures within a sample from the search space. Early works in prediction-based NAS revolve around learning curve extrapolation. Learning curve extrapolation can assist neural architecture search (NAS) and hyperparameter optimization (HPO) terminates the training process early, thus, reducing the computational cost. Klein et al. \cite{learningcurve} propose a probabilistic approach to modeling training curves. Their method is similar to \cite{sppedup}, which used basis functions to extrapolate learning curves from the architecture performance observed via the training process. The differences lie in the usage of the Bayesian neural network allowing their method can be hyperparameter-aware. This allows the sharing of knowledge between different learning curves, which is more data efficient than the previous method \cite{sppedup}. They conduct experiments on small and large convolutional neural networks on CIFAR datasets \cite{cifar}. They surpassed previous methods on HPO and even show strong correlation between predicted accuracy and accuracy of unobserved learning curves. The above method can also be used in NAS if hyperparameters are architecture-related, such as the number of layers, the width of $s^{th}$ layer, etc. Kandasamy et al. \cite{nasbot} leverage GP as a surrogate model to predict the performance of architectures and employs Bayesian optimization to guide the search process efficiently. Their Bayesian procedure incorporates optimal transport based (pseudo-)distance for neural network graphs (OTMANN) to define the kernel function of architectures. With the assistance of GP as a surrogate model, NASBOT finds better MLP architectures for the regression problems: blog feedback \cite{blog}, indoor location \cite{indoor}, slice localisation \cite{slice}, naval propulsion \cite{naval}, protein tertiary structure \cite{protein}, news popularity \cite{news} and CNN architecture for the classification task: CIFAR-10 \cite{cifar}, as compared to TreeBO, EA, RAND. While they deploy in more general spaces and consume less computational cost, NASBOT shows lower performance than prior work \cite{ProgressiveNAS, HierarchicalNAS, rl1}. Advancements in neural graph processing techniques like Graph Convolutional Networks \cite{gnn} have replaced complicated distance measurement methods for Bayesian optimization. Dudziak et al.'s work \cite{BRP-NAS} is one of the early adopters of this technique in Neural Architecture Search. They introduced transfer learning from latency prediction task to assist accuracy prediction task. Note that, latency measurement is much cheaper than accuracy measurement, and deep learning scales with data; therefore, they can re-use the feature trained from a much larger dataset to solve accuracy prediction with limited data. Applying transfer learning robustifies their predictor's performance from 0.834 to 0.890, calculated using Spearman correlation between predicted score and test-accuracy in NAS-Bench-201 \cite{nb201}. In addition, their paper suggests an iterative method for choosing data. This method, called iterative data selection, aims to rank the top-K architectures instead of the low-performing ones. By using iterative data selection, their method performs better in an End-to-End NAS procedure in NAS-Bench-201 \cite{nb201} and NAS-Bench-101 \cite{nb101} benchmarks. Moreover, they successfully discovered $97.6\%$ test-accuracy on CIFAR-10 \cite{cifar} within the DARTS search space \cite{DARTS}. Lu et al. \cite{TNASP} used a transformer instead of GCNs or MLP for structural encoding. Their main contribution lies in the method to process architecture graphs using a transformer, they use Laplacian matrix-based positional encoding. The Laplacian graph is transformed using a multi-layer perceptron. Then, it is combined with operational encoder to create positional encoding for the transformer. Their self-evolution uses historical validation accuracy to optimize their NAS procedure to guide the predictor to avoid overfitting. Their method achieved good results on both NAS-Benchmarks and $97.48\%$ on DARTS search space \cite{DARTS}.

\section{Proposed Method} \label{sec:method}

Our proposed method has three main components: building the neural performance predictor, optimizing this model, and using it in an end-to-end NAS process.

\begin{itemize}
    \item The construction of the neural performance predictor is centered on a representation mechanism, which is detailed in Section \ref{sec:representation_mechanism}. The specifics of representing neural network components are expanded upon in Section \ref{sec:conv}, with an emphasis on the Fourier-Transform-based representation for convolutional layers. Section \ref{sec:scorer} explains how we use extracted topological features to compute architectural scores.

    \item For model optimization, we introduce an optimal-transport-based method. This, combined with an iterative method, lets us make the most of multiple architectural datasets available. Details can be found in Section \ref{sec:model_opt}.

    \item As for the search algorithm, we employ three widely used evolutionary algorithm (EA) strategies, designed for discrete (simple genetic algorithm), continuous (differential evolution \cite{DE}), and multi-objective (NSGA \cite{NSGA}) scenarios. This algorithm modifies a constrained single-objective problem into a multi-objective one. This change allows for efficient identification of top-performing architectures within specific constraints. This aspect of our end-to-end NAS process is laid out in Section \ref{sec:searchalgo}.
\end{itemize}

\subsection{Representation Mechanism} \label{sec:representation_mechanism}
In prior studies, Zero-Shot NAS centered on encapsulating architectural operations in a manner that represents the capabilities of those operations. For instance, Zen-NAS \cite{Zen-NAS} leverages the log-sum of finite differentials of a neural network (based on expected Gaussian complexity) and the average variance of BN layers as functional complexity. Conversely, our representation mechanism primarily aims at producing a distinctive encoding for each architecture, thereby empowering the neural network to discriminate efficient architectures from inefficient ones. Our design principle stipulates that identical architectures should yield identical representations. For example, the outcome of stacking two consecutive batch normalization layers, ReLU activations, or identity layers should be identical. The representation design should also resonate with what it symbolizes. For example, under our design principle, a ReLU activation or a convolutional operation with set weights are appropriate representations for ReLU activation and convolutional layer, respectively. In contrast, using a one-hot vector or a multi-layer perceptron to depict an activation function or convolutional operation does not fit our design approach.

\begin{figure*}
        \centering
        \includegraphics[scale=0.39]{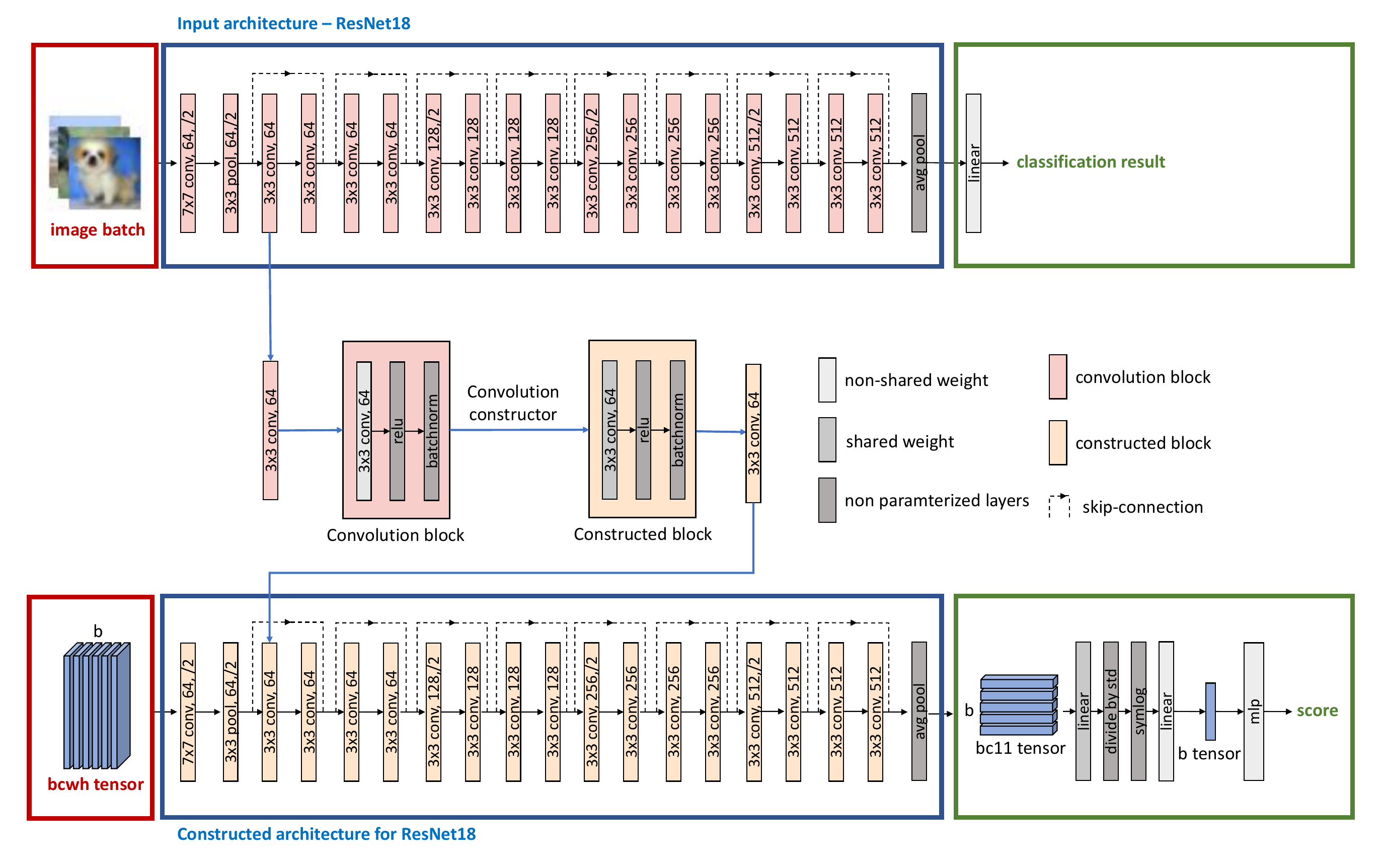}
        \caption{The process of using the representation mechanism to extract topological features of the input architecture and compute scores using MLP. Convolution layers and other components are replaced by their respective representations to form constructed blocks. They are the representation of the convolution blocks and are the building blocks for the representation of the architecture which is called constructed architecture. Red, blue, and green sections represent the pairing of neural network computation components and their corresponding representations.}
        \label{fig:wholeprocess}
    \end{figure*}

To realize this, we decompose the computation process of the neural network into three parts: the input tensor, the architectural components, and the loss computation. Figure \ref{fig:wholeprocess} describes the decomposition of ResNet-18 \cite{ResNet} computation as well as the replacement of its components. The input tensor is replaced with a learnable tensor mirroring the size of an image batch, symbolizing the dataset that the architecture is trained on. The architectural components are replaced with their respective representations, acting as topological information extractors. The loss function is symbolized by a multi-layer perceptron that evaluates the performance of different neural network architectures, thereby creating a mapping from the embedded topological structure to an architecture's performance on a particular task. The following section provides further details on how we represent the components of a neural architecture.

\begin{figure*}
        \centering
        
        \subfloat[Neural-based Dimensional Reduction]{%
            \includegraphics[width=0.48\textwidth]{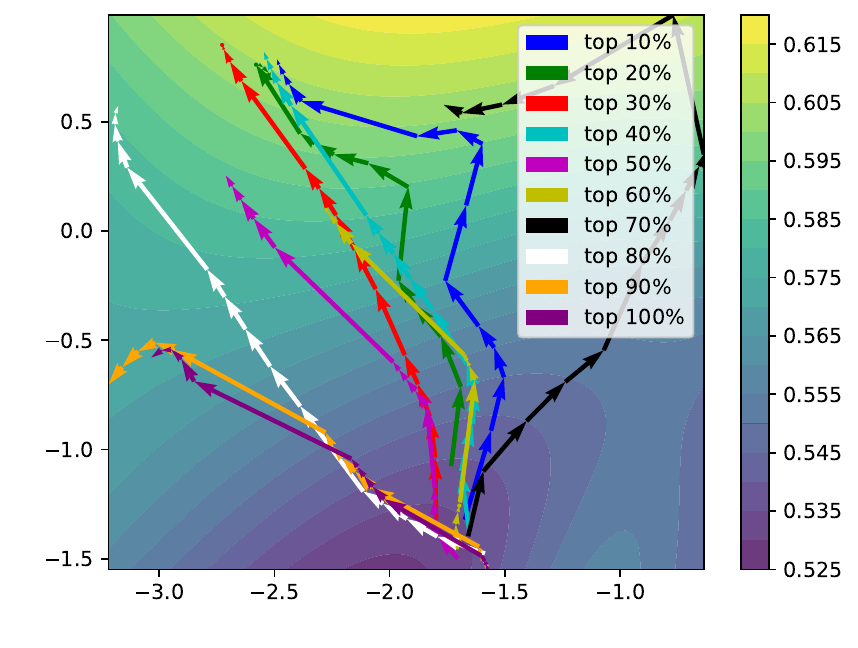}%
            \label{fig:neural}%
        }%
        \hfill
        \subfloat[PCA-based Dimensional Reduction]{%
            \includegraphics[width=0.48\textwidth]{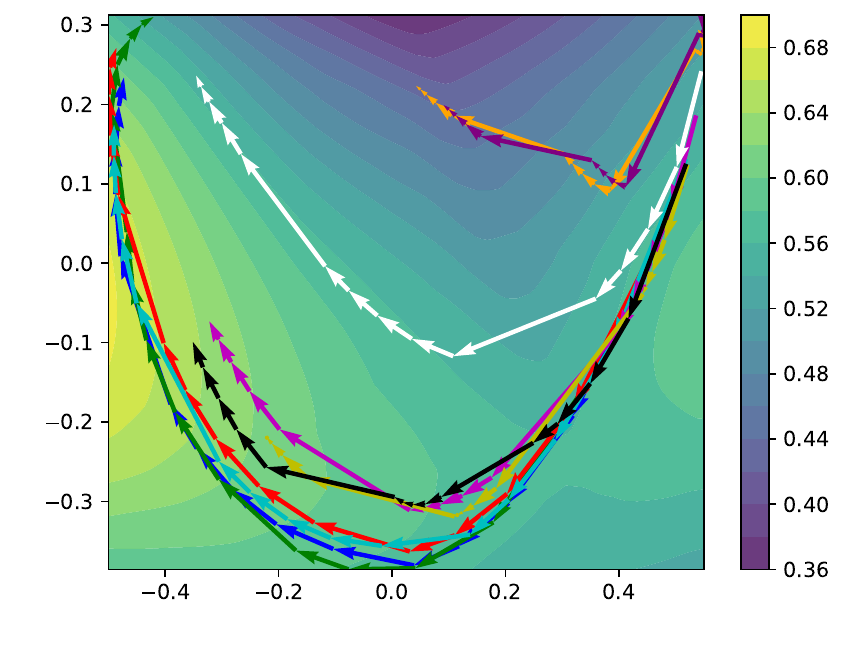}%
            \label{fig:pca}%
        }%
        
        \caption{Score-Representation Transition of NAS-Bench-201 Architecture Layers. Quivers show the transitions of representation across NAS-Bench-201 15 blocks. Heatmaps show the landscape of the scorer obtained via Neural Networks (for accurate score landscape) or Reverse-PCA (for informative representation distance). Architectures with higher scores indicate their superior expected performance. The figure aims to show that top-half architectures tend to perform better when it gets deeper and bottom-half architectures' performance saturates at certain depth.}
        \label{fig:score_rep_transistion}
    \end{figure*}

\begin{figure}
        \centering
        
        \includegraphics[width=0.6\linewidth]{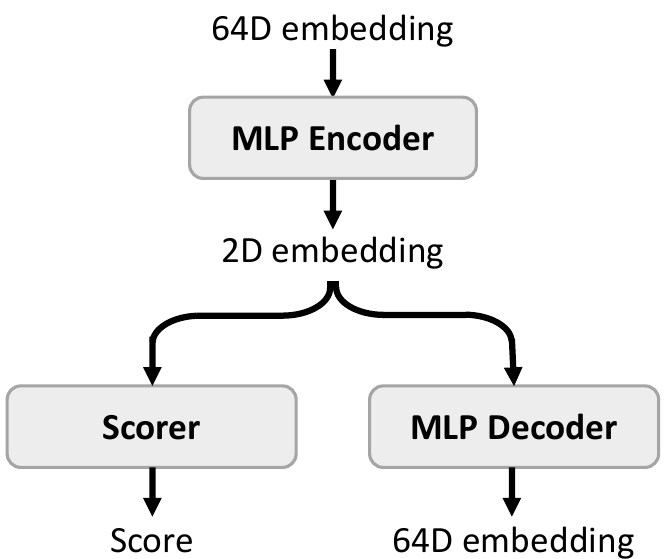}
        \caption{Dimension Reduction via Neural Auto-Encoder}
        \label{fig:neural_dim_reductor}

    \end{figure}

In simple terms, each layer representation slightly transforms its inputs, causing changes in the neural network architecture encoding. The scorer then evaluates these structural changes by analyzing the encoding. Figure \ref{fig:score_rep_transistion} illustrates how different scores result from changes in each layer. The heatmap depicts the scores of architectures on a 2D projection of the architecture encoding. Figure~\ref{fig:score_rep_transistion} exhibits that the \textbf{scores of top architectures tend to increase as the architectures get deeper}. To create this figure, we first randomly select 200 architectures from NAS-Bench-201. We then compute the encoding of architectures using an algorithm we'll describe in section Section \ref{sec:conv} and Section \ref{sec:scorer}. We reduce the dimensions of these encoding vectors to 2, using either a neural network or PCA. For the neural network approach to dimension reduction, we use an autoencoder-decoder\&scorer model, as shown in Figure \ref{fig:neural_dim_reductor}. The model combines an $L_2$ construction loss for the encoder-decoder pair and an MSE loss for the encoder-scorer pair. For the PCA dimension-reduction version, we train the PCA on 200 architecture encodings and then compute the score using the scorer on the reverse-PCA of the 2D-space. These two methods help us construct the heatmap of the graph. The score-representation transition is derived from the output of 15 cells of 10 NAS-Bench-201 architectures. We randomly select architectures from the top 0-10\%, 10-20\%, ..., 90-100\% performance brackets (measured using test-accuracy). We then use a global average pooling on the feature maps to create encoding vectors. The encoding goes through either a neural-based dimensional reduction or a PCA derived from training 200 architectures. The neural-based map helps in visualizing accurate projections from 2D variables to the score, while the PCA-based map is useful for visualizing the magnitude and direction of encoding changes through the layers.

    \subsection{Neural Network Representation} \label{sec:conv}
    
    \subsubsection{Convolutional Representation}
    
    Work such as BRP-NAS \cite{BRP-NAS} applied Graph Convolutional Networks (GCNs) to prediction-based NAS, demonstrating that layered linear transformations can effectively characterize different convolutional layers within a neural network architecture. Therefore, a single shared convolutional kernel should be able to represent convolutional layers across different depths. This approach ensures a fixed output representation for each convolutional layer by avoiding the random output that comes from random initialization. Random outputs can be solved using set processing techniques \cite{Deepsets}. However, its demand for significant computational resources hinders us from trying this approach. Thus, we opted to use the shared convolutional kernel method for non-random outputs.
    
    The shape of a convolutional kernel depends on the settings of the convolutional layer, including channel inputs, channel outputs, and padding settings. This makes it impractical to simply learn a weight matrix. Therefore, we need to create a method that can construct weight matrices based on configurations. We employed Fourier Transform (FT) as a method to construct weight matrices, specifically the magnitude of the Fourier Transform with an orthonormal basis. To adjust for the channel configuration and kernel size, we use zero-padding and trimming before applying the Discrete Fourier Transform (DFT), which are standard signal processing techniques.
    
    Let's denote $X[k](n=1,2,\dots, K)$ as the sequence Discrete Fourier transformed from the sequence $x[n](n=1,2,\dots, N)$. The mean and variance of the DFT coefficients can be determined as follows:
    
    \begin{equation} \label{eq:expected_fourier}
    \mathbb E\left[X[k]\right] = \min{\left(1, \sqrt{\frac{N}{K}}\right)} \mathbb E\left[x[n]\right]
    \end{equation}
    
    \begin{equation} \label{eq:variance_fourier}
    \mathbb V\left[X[k]\right] = \min{\left(1, \frac{N}{K}\right)} \mathbb V\left[x[n]\right]
    \end{equation}
    
    Equations \ref{eq:expected_fourier} and \ref{eq:variance_fourier} show that using zero-padding changes the mean and the variance in DFT coefficients by factors of $\sqrt{\dfrac{N}{K}}$ and $\dfrac{N}{K}$, respectively. Therefore, if $N < K$, we divide the weight matrix by $\sqrt{\dfrac{N}{K}}$ to preserve the mean and variance.
    
    We store the frequency of the convolutional kernel as $n_{in} \times n_{out}$ kernel maps. To adjust the size of these kernel maps and generate the desired number of kernel maps, we use a 2D DFT and a 1D DFT, respectively. This process uses Fast Fourier Transforms (FFTs) with either zero-padding or trimming.

    \subsubsection{Variance Unitization}
        We distinguish the term \emph{variance unitization}, which refers to the division by standard deviation, from \emph{normalization}, which subtracts the mean before dividing by the standard deviation.
        
        Kaiming He initialization is a method designed to preserve the variance of a layer's output after applying a ReLU activation function. It initializes the weight matrix using a normal distribution with mean 0 and variance $\sqrt{\dfrac{2}{n}}$, where $n$ is the number of input channels of the weight matrix. Similarly, after applying our convolutional representation, we divide the output by $\sqrt{\dfrac{2}{n}}$, where $n$ is the number of input channels. We refer to $\sqrt{\dfrac{2}{n}}$ as the \emph{unitization factor} in this context.
        
        However, with the presence of multi-branch models and residual links, simply dividing the output by $\sqrt{\dfrac{2}{n}}$ does not ensure the preservation of variance. To address this issue, we propose a v-norm algorithm (variable-normalization) that calculates the unitization factor for each convolutional layer. The v-norm algorithm breaks the computation of the neural network representation into two forward passes: one for computing the unitization factor and the other for computing scores. We split the process into two forward passes because we do not want to pass or store the gradient of the sampling process. Note that while the model is used to test/search, computing two forward passes is not necessary since we do not need gradients. When computing the unitization factor for each convolution representation layer, the algorithm performs the following steps:
        
        \begin{enumerate}
        \item Receive input or output from previous layers.
        \item Compute the convolution output using the shared initial convolutional kernel.
        \item Store the standard deviation of the output as the unitization factor.
        \item Divide the computed output by the unitization factor.
        \item Feed the output to the next layers.
        \end{enumerate}

        The v-norm algorithm helps prevent numerical overflow when the model convolves using one single kernel, as compared to methods based on Kaiming He initialization. This makes the model more amenable to optimization. However, unitization could result in some loss of information regarding the model size, which could potentially affect the model's performance.

        \subsubsection{Non-Convolutional Representation}
            We here devise straightforward methods to represent batch normalization layers, pooling layers, and activation functions. Average pooling, max pooling, and ReLU activation functions are represented in their original forms, without transformation or encoding. We handle batch normalization by normalizing across the batch dimension and appending an extra batch dimension to the input-like tensor. Although our work does not currently extend to developing representations for dropout and drop-connect, we anticipate that our methodology should generalize well to accommodate these components in the future.

    \subsection{Scorer Construction} \label{sec:scorer}

        The computation of our neural performance indicator involves three sets of parameters: input-like tensor $I$, shared convolutional weight $W$, and the weights of the multi-layer perceptron (MLP). The feature extraction from the neural network, $\mathrm{NN}$, is performed by replacing its operators with their respective representations using the shared convolutional weight $W$.
        
        Given that our batch normalization introduces an additional batch dimension to the input-like tensor, we incorporate two channel-wise linear transformations and a batch-wise MLP ($\mathrm{MLP}_b$). The first linear transformation, $\mathrm L_1$, possesses a variable number of input channels and a fixed number of output channels. It employs the shared convolutional weight since this enables the extraction of valuable information from the weight, even with limited data. We formulate a 1x1 convolutional kernel for the $L_1$ linear transformation.
        
        The SymLog activation function is utilized between the two linear transformations to handle the variation in tensor scales across different architectures and architectural spaces. This choice is inspired by the work of  Hafner et al. \cite{hafner2023dreamerv3}.
        
        The second linear transformation, $L_2$, takes the output of $\mathrm{SymLog}(L_1)$ as input and has a single output channel. The output of $L_2$ is transposed (i.e., swapping the batch and channel dimensions and removing the channel dimension) before being fed into $\mathrm{MLP}_b$. The final layer of $\mathrm{MLP}_b$ is a linear transformation with an output size of 1, providing the score of the architecture.
        
        For variants using the v-norm algorithm, an additional layer is added to the scoring process to unitize the variance of the $L_1$ linear transformation output. While the usage of v-norm algorithm combined with unitization after $L_1$ makes the usage of SymLog activation function useless, we decide to keep it for the ease of engineering, paving the way for future combinations of topological extractors of different kinds. The entire process is depicted in Figure \ref{fig:wholeprocess}, and the SymLog function is defined in Equation \ref{eq:symlog}.
        
        \begin{equation}
        \mathrm{SymLog}(x) = \mathrm{sign}(x)\log\left(\left|x\right| + 1\right) \label{eq:symlog}
        \end{equation}

\subsection{Optimization of Model} \label{sec:model_opt}

        \subsubsection{Optimization in a Single Space} 
            The previous prediction-based NAS (BRP-NAS) \cite{BRP-NAS}\ used a binary relation predicting method. They found that learning score is easier than predicting accuracy. The binary relation predicting method uses $O(n^2)$ binary relationships from $n$ architecture-accuracy pairs. However, this method gives equal weight to all binary relation data points, which can be a limitation. Instead, we recommend using a fast differentiable ranking method \cite{fast-soft-sort} to directly optimize the scoring-based model for Spearman correlation. The mentioned method utilized projections onto a convex hull of permutations to make differentiable sorting operators.

            Conventionally, deep learning models are trained on GPUs because of the benefit of parallelization, fastening training process up to three times. However, the fast-differentiable-ranking-based algorithm consumes $O(s)$ times more memory to train models than the binary relation algorithm. This makes optimization on GPUs consume tremendous resources; therefore, we resort to CPU-based training. The following shows that optimization on CPUs using fast differentiable ranking method can be as effective as optimization on GPUs using binary relation method. The fast-differentiable-ranking-based algorithm can optimize $O\left(s^2\right)$ binary relationships simultaneously in $O(s \log s)$ cost compared to binary relation algorithm's $O\left(s^2\right)$ cost. This makes the algorithm $O\left(\dfrac{s}{\log s}\right)$ more efficient than the binary relationship algorithm. With sample size of 64, the algorithm is likely to run on CPUs as efficiently as the binary relationship running on GPUs.
            
 
        \subsubsection{Optimization in Multiple Spaces} \label{sec:mulspace}
             Differences can be found in the collected search spaces (see in Table \ref{tab:training_settings}), mainly due to variations in the training process, the optimization algorithms used, and the chosen hyperparameters. For instance, one dataset might contain architectures trained for one epoch, while another might contain architectures trained over 2,000 epochs. These differences can distract the neural network from learning useful architecture design biases.
            
            We propose two solutions to reduce the differences in different training settings based on the assumption that good architectures perform better than bad ones across various training settings. These solutions are:
            
            \begin{itemize}
            \item Updating model parameters by iterating over datasets and using samples from each dataset.
            \item Computing a sample from each dataset, then using gradient accumulation to update model parameters collectively.
            \end{itemize}
            
            We prefer the first option in our research, as it uses less computational resources, despite being more unstable.
            
            Alternatively, we suggest an ensemble approach incorporating all performance predictors. We combine eight of our models trained across eight different datasets using a linear combination of normalized sigmoid of scores, as shown in Equation \ref{eq:norm_sig}. The normalized sigmoid addresses the out-of-distribution problem often seen in predictors trained on a single dataset. Our aim is to average the Spearman correlation across datasets. As the objective function is non-differentiable, we use differential evolution to find these linear coefficients within a bound of (0, 1). However, any other black-box optimization algorithm should work well.
    
            \begin{equation} \label{eq:norm_sig}
            f(x) = \sum_{i=1}^{8} w_i \cdot S\left(\frac{s_i(x) - \mu_i}{\sigma_i}\right)
            \end{equation}
            
            Here, $f(x)$ is the output of the linear combination, $s_i$ are the scorers trained on $i^{th}$ search space, $w_i$ are the corresponding weights, $\mu_i$ and $\sigma_i$ are the mean and standard deviation of $s_i(y)$ for architectures $y$ in search space $i$, and $S$ is the sigmoid function defined as:
            
            \begin{equation}
            S(x) = \frac{1}{1 + e^{-x}}
            \end{equation}
            
            The ensemble method offers the advantage of parallelization, allowing for simultaneous training across all search spaces. But it requires longer inference times, needing eight forward passes. With more topological information from the neural network computational graph, we expect the ensemble method to perform better than the dataset iteration approach.

    \subsection{Search Algorithm} \label{sec:searchalgo}
        Our predictive model, equipped with our v-norm algorithm for variance stabilization, is not easily affected by changes in model capacity. This technique mirrors how information changes and flows within the model. We've designed an algorithm that combines insights from model capacity and topological structure. We chose the number of parameters as a measure of model capacity, for simplicity and efficiency.
        
        \subsubsection{Objective-Space Ensemble}
            We bring these components together by projecting the scores from our predictor onto a two-dimensional objective space. One axis represents the performance score from our predictor, while the other represents capacity sensitivity, measured by number of parameters. This setup turns the task into a multi-objective optimization problem. The goal is to find a balance between the performance score and architectural capacity. This alleviates the weakness of v-norm algorithm, which destroys information regarding the parameter count.
        
        \subsubsection{Optimization Algorithm}
            To solve this multi-objective optimization problem, we use the Non-dominated Sorting Genetic Algorithm II (NSGA-II) \cite{NSGA}. NSGA-II is known for its rank-based fitness evaluation, which can handle differences in scale between the performance score and number of parameters. We also include Differential Evolution (DE) operations \cite{DE} for the crossover and mutation processes. This combination is especially good at handling ordered elements, like the number of output channels. With DE operations, NSGA-II can effectively explore the search space while considering multiple objectives. By keeping the diversity among solutions, it can find high-performance models within an upper bound of the number of parameters.

            The iterative operation can be explained detailedly as below:

            \begin{itemize}
                \item Step 1: Create initial population with evolutionary hyperparameters: population size and variable bounds.
                \item Step 2: Evaluate the fitness of individuals using two criteria: the neural score and the number of parameters (bigger is better, both).
                \item Step 3: Perform non-dominated sort, then calculate the crowding distance of individuals to perform NSGA selection.
                \item Step 4: Creating offspring from population via simple genetic algorithm (for discrete variables) and differential evolution operations (for ordered variables). Since differential evolution operates on continuous space, it returns continuous values. To apply to our problem, the continuous values are rounded.
                \item Step 5: If not reached the maximum number of iterations, go to Step 2. Otherwise, return the neural network architecture with the highest score satisfying certain lower bound of number of parameters.
            \end{itemize}
        
\section{Experiments} \label{sec:experiment}
    %
    The datasets used in the experiments were NAS-Bench-201 \cite{nb201}, NAS-Bench-101 \cite{nb101}, NAS-Bench-Macro \cite{macro}, NDS-Amoeba, NDS-DARTS, NDS-NASNet, NDS-ENAS, NDS-PNAS \cite{nds}. We also searched for architecture under 1 million parameters in the ResNet-like search space Zen-NAS \cite{Zen-NAS} offered and used their results of performance in End-to-End NAS Procedure of performance predictors for comparison. The neural \textbf{net}work architectures found by the neural \textbf{net}work performance evaluators are called \textbf{Net$^{2}$}. The scorers using fourier transform based features are called \textbf{FT-ScoreNet}s. All the scorers are trained via Adam optimizers with hyperparameters: $lr = 0.001$, $\beta_1 = 0.9$, $\beta_2 = 0.95$. The input-like tensors for CIFAR-10 and ImageNet dataset are (64, 32, 32, 3) and (64, 224, 224, 3), respectively. The size of frequency matrices is (64, 64, $k$, $k$), where $k$ refers to the kernel size of the largest convolutional operation existed in the dataset.

    \subsection{As a Zero-Shot NAS}
        We conduct experiments on optimizing for score-accuracy correlation. The model is trained on NAS-Bench-201 (CIFAR-100) \cite{nb201}, NAS-Bench-101 \cite{nb101}, NAS-Bench-Macro \cite{macro}, NDS-DARTS, NDS-NASNet, NDS-Amoeba, NDS-ENAS, NDS-PNAS \cite{nds}. It is then tested on these same datasets, each was tested using 1,000 samples. We also train the model on every dataset combined and test on Zen-NAS-residual-like search space. We compute score-accuracy correlation of Zen-Score in those datasets. We compare our results against Zen-Score, NASWOT-Score, and performance predictors in \cite{LightweightNAS}. The regularization strength of 3.0 is used for the differentiable sorting algorithm. Models in NAS-Bench-201 and NAS-Bench-Macro are trained using a sample of 64 architectures. Since the models in NDS and NAS-Bench-101 are bigger, sample size of 7 is used to train the scorer. The numbers of training steps for NAS-Bench-201, NAS-Bench-101, NAS-Bench-Macro, and NDS are 496 (2 epochs), 1,440 (0.02 epochs), 208 (2 epochs), 1,440 (2 epochs).

        \subsubsection{Score-Accuracy Correlation}
            In this experiment, we show that neural-based method can outperform human experts in measuring score-accuracy correlation. Table \ref{tab:score_accuracy_correlation} and Table \ref{tab:score_accuracy_correlation_not_vnorm} describe score-accuracy correlations for various models using the v-norm algorithm and not using the v-norm algorithm, respectively. They are evaluated using Spearman's rank coefficient. Each cell (column $i$, row $j$) indicates score-accuracy correlation of a model trained on dataset i and tested on dataset j. Evaluations were performed on a random sample of 1,000 architectures from each search space (20\% of NDS). Zen-Score, NASWOT-score, and number of parameters correlations with test-accuracy on benchmarks are measured by us. \textbf{Bold} indicates neural predictors outperforming all handcrafted ones, and \underline{underline} indicates outperformance over two handcrafted predictors. Unmarked cells represent neural models tested and trained on identical datasets.

\begin{figure*}
            \centering
            
            \subfloat[DARTS]{%
                \includegraphics[width=0.3\textwidth]{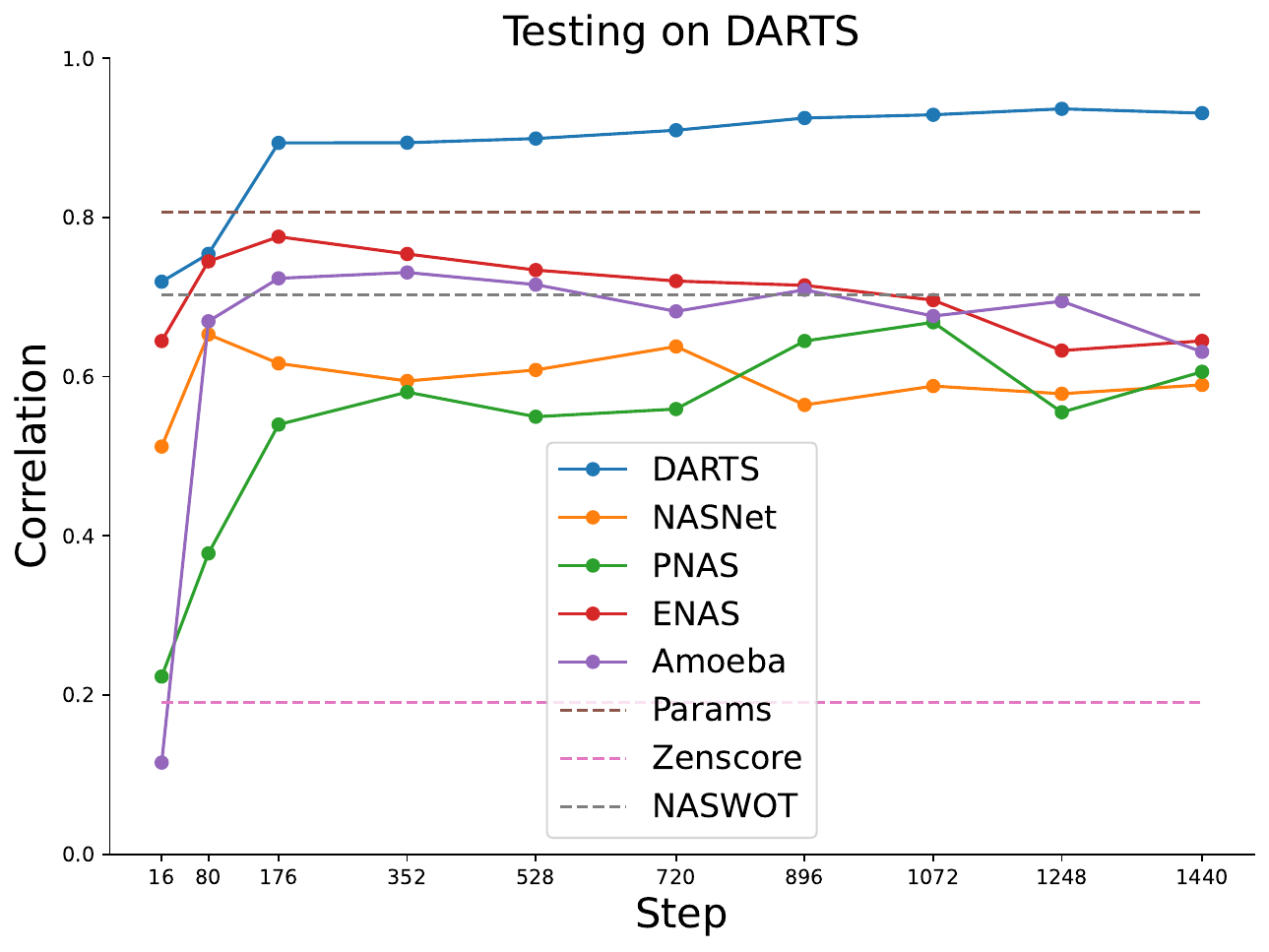}%
                \label{fig:learning_curve1}%
            }%
            \hfill
            \subfloat[NASNet]{%
                \includegraphics[width=0.3\textwidth]{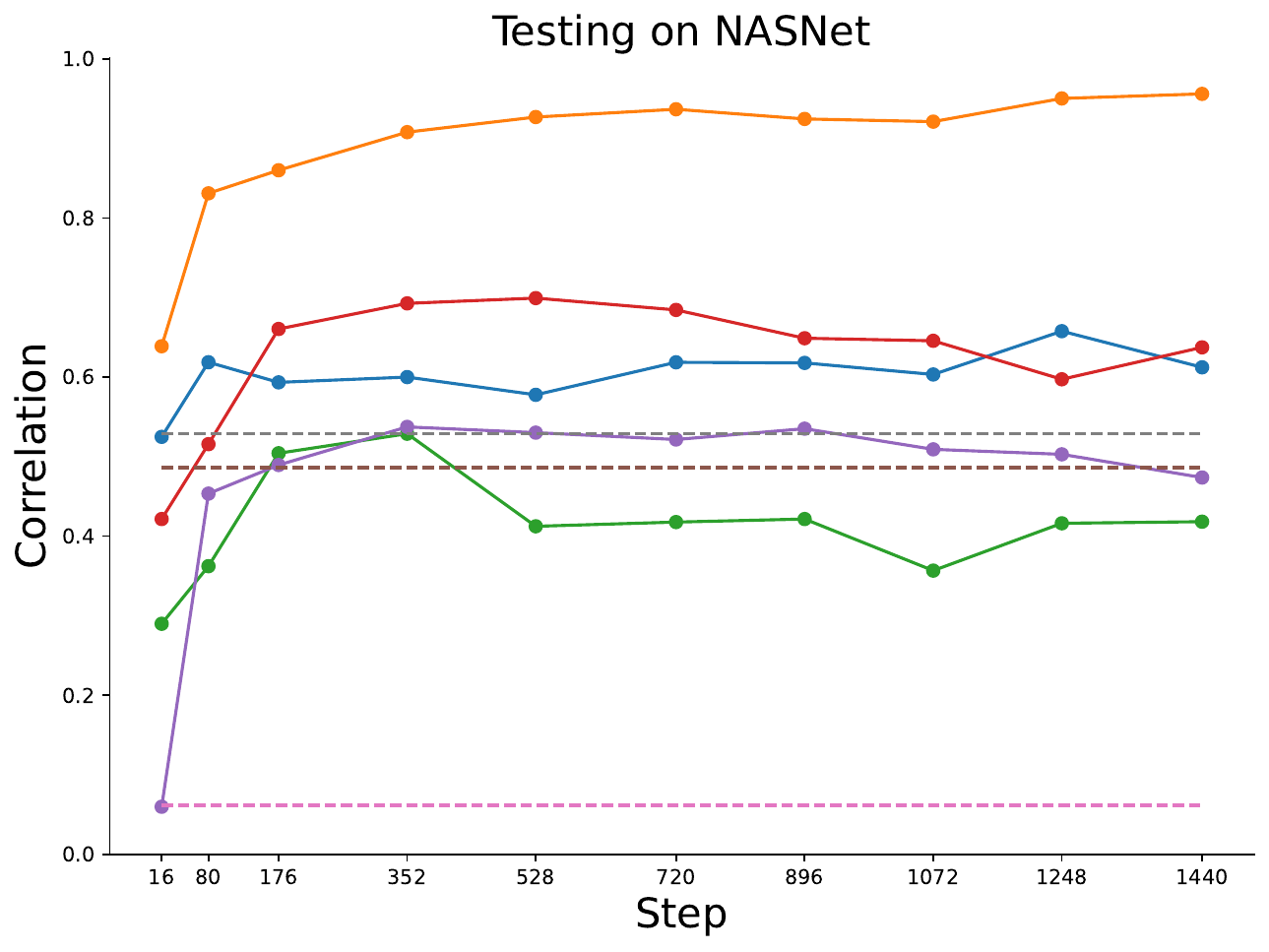}%
                \label{fig:learning_curve2}%
            }%
            \hfill
            \subfloat[Amoeba]{%
                \includegraphics[width=0.3\textwidth]{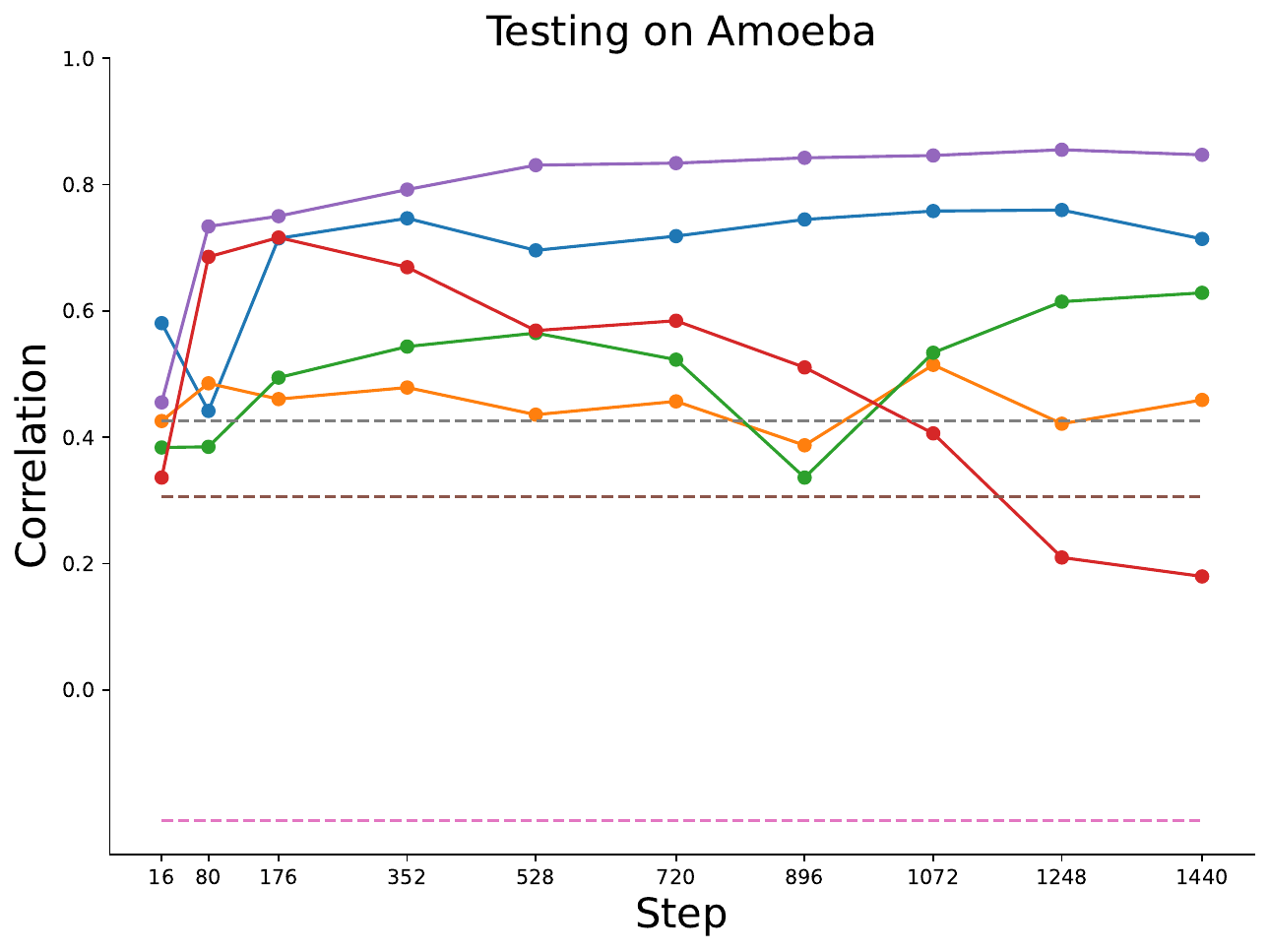}%
                \label{fig:learning_curve3}%
            }
            
            \vspace{1cm} 
            
            \subfloat[ENAS]{%
                \includegraphics[width=0.3\textwidth]{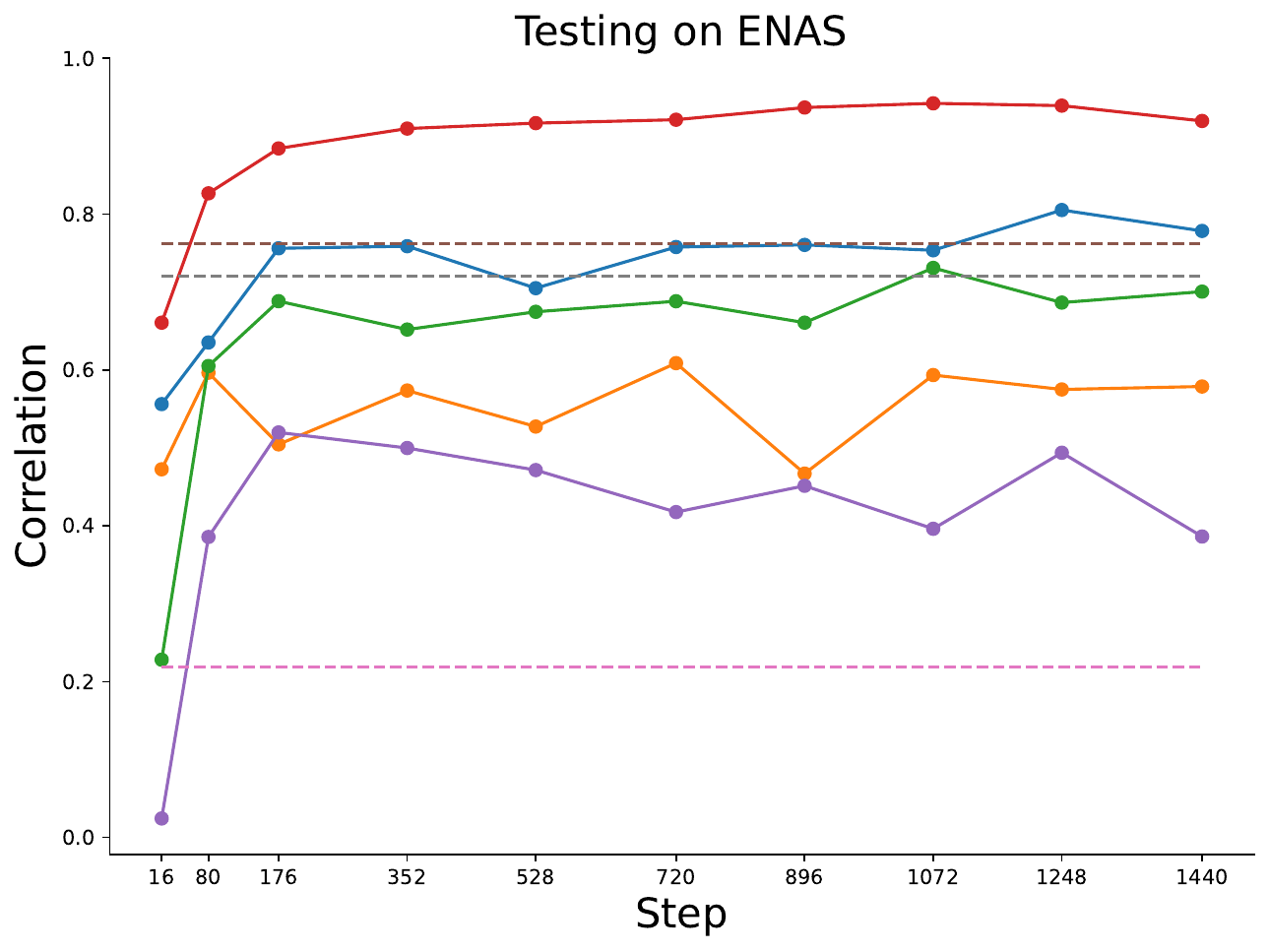}%
                \label{fig:learning_curve4}%
            }%
            \hfill
            \subfloat[PNAS]{%
                \includegraphics[width=0.3\textwidth]{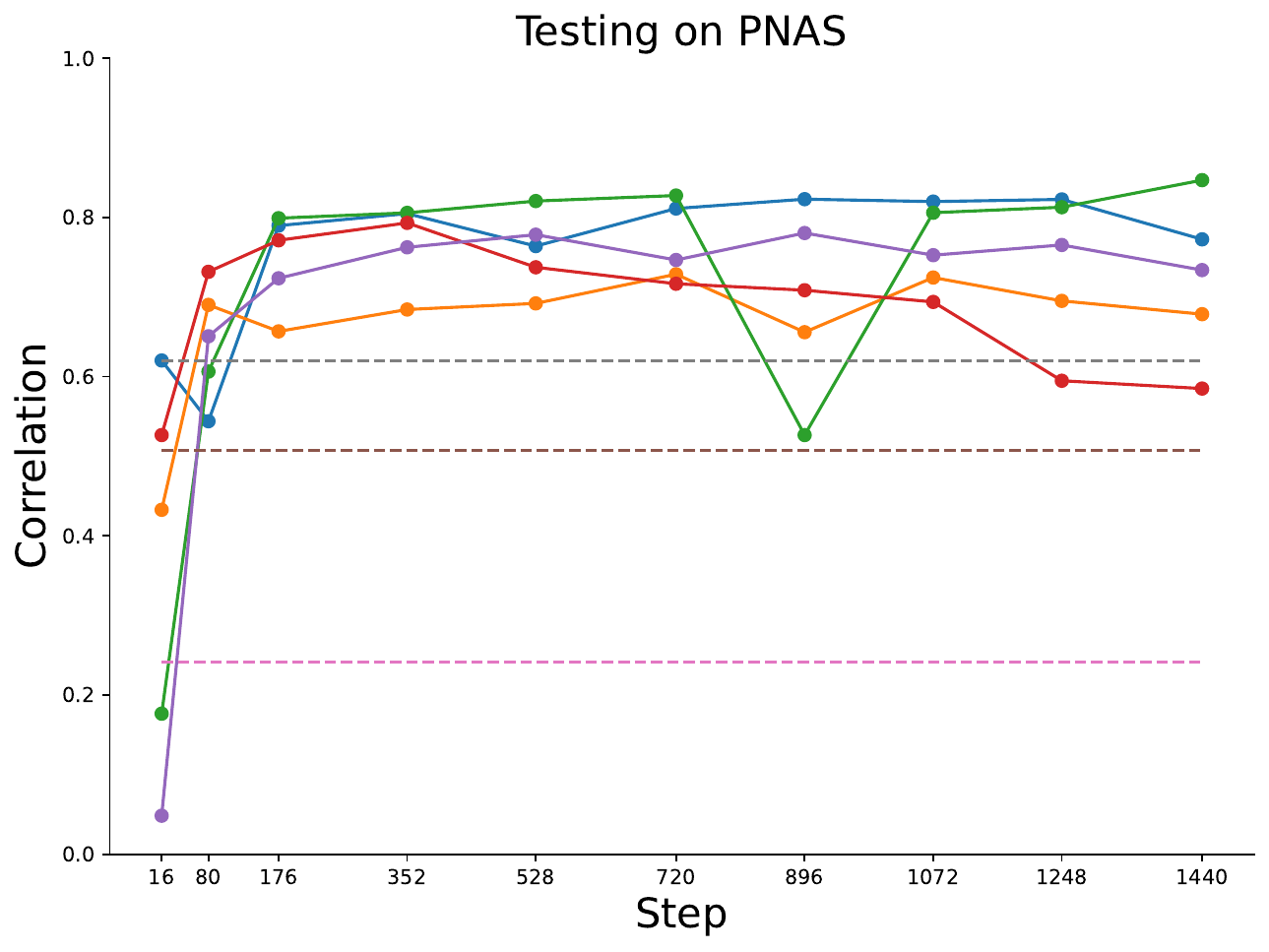}%
                \label{fig:learning_curve5}%
            }%
            \hfill
            \subfloat[NAS-Bench-201]{%
                \includegraphics[width=0.3\textwidth]{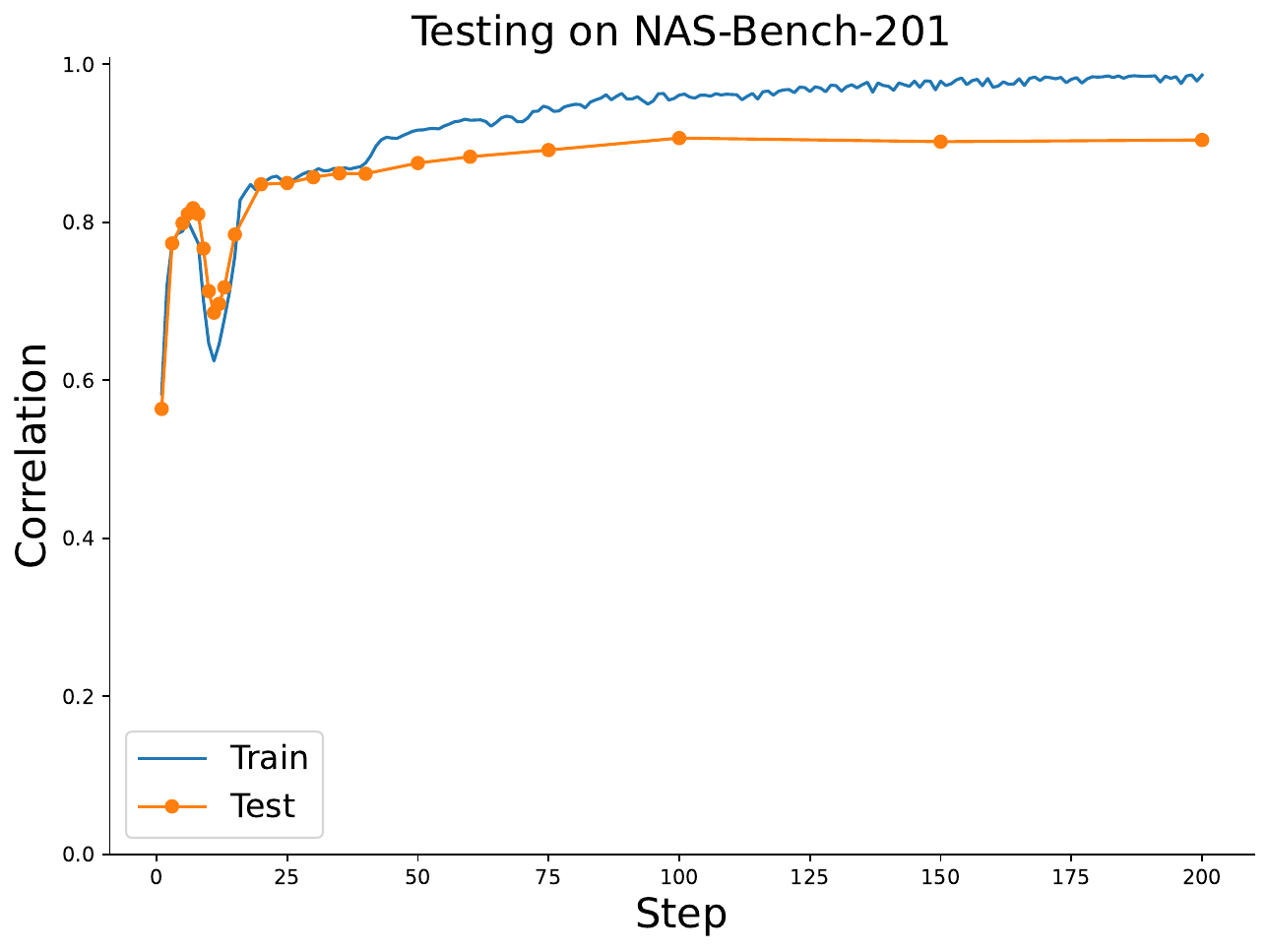}%
                \label{fig:learning_curve6}%
            }
            
             \caption{Learning curve of neural scorers on data-scarce settings: Zero-Shot NAS settings on ImageNet (\ref{fig:learning_curve1}-\ref{fig:learning_curve5}) and Prediction-based NAS settings on CIFAR-100 (\ref{fig:learning_curve6}).}
            \label{fig:learning_curve}
        \end{figure*}

            Figure \ref{fig:learning_curve1}-\ref{fig:learning_curve5} shows the learning curve of models in trained and tested on each other within NDS-ImageNet datasets. Since NDS-ImageNet datasets only have around 120 architectures each, we refer to this setting as one of the data-scarce settings. They are trained using 1440 steps and the efficiency of each performance predictor is measured in Spearman correlation.

            \begin{table*}
            \caption{Score-accuracy correlations for architecture performance evaluators with v-norm on CIFAR datasets}
            \label{tab:score_accuracy_correlation}
            \centering
            \begin{tabular}{c|rrrrrrrr|rrr}
            \toprule
                &\multicolumn{8}{c|}{FT-ScoreNets are trained on datasets \cite{nds, nb201, nb101, macro}}& \multicolumn{3}{c}{Handcrafted} \\
                \cmidrule(r){2-9} \cmidrule(r){10-12}
                & DARTS & NASNet & Amoeba & ENAS & PNAS & NB201 & NB101 & Macro & Zen-Score \cite{Zen-NAS} & Params & NASWOT \cite{naswot}\\
                \midrule
                DARTS & 0.746 & 0.560 & \textbf{0.758} & 0.589 & \textbf{0.709}& 0.140 & -0.137 & 0.281 & 0.448 & 0.668 & 0.647\\
                NASNet & \textbf{0.633} & 0.803 & \textbf{0.699} & \textbf{0.713} & \textbf{0.648} & \textbf{0.442} & 0.199 & \textbf{0.453} & 0.102 & 0.411 & 0.418 \\
                Amoeba & \textbf{0.709} & \textbf{0.662} & 0.797 & \textbf{0.704} & \textbf{0.755} &0.335 & 0.220 & 0.289& -0.046 & 0.343 & 0.276\\
                ENAS & \textbf{0.735} & \textbf{0.721} & \textbf{0.760} & 0.777 & \textbf{0.642} &  0.425 & 0.033 & 0.375&  0.232 & 0.561 & 0.532\\
                PNAS & \textbf{0.682} & \textbf{0.580} & \textbf{0.758} & \textbf{0.696} & 0.723 & 0.029 & -0.091 & 0.116 & 0.259 & 0.541 & 0.496\\
                NB201 & 0.695 & 0.390 & \underline{0.789} & \underline{0.756} & \underline{0.789} & 0.939 & 0.704 & 0.721 &  0.429 & 0.725 & 0.824\\
                NB101 & \textbf{0.636} & \underline{0.560} & \underline{0.529} & \underline{0.494} & 0.341 & 0.340 & 0.877 & \underline{0.463} & 0.628 & 0.431 & 0.388\\
                Macro & \underline{0.710} & 0.650 & 0.540 & 0.591 & 0.598 & \underline{0.754} & 0.312 & 0.979& 0.682 & 0.317 & 0.900 \\
            \bottomrule
          \end{tabular}
          \end{table*}

          \begin{table*}
            \caption{Score-accuracy correlations for architecture performance evaluators without v-norm on CIFAR datasets}
            \label{tab:score_accuracy_correlation_not_vnorm}
            \centering
            \begin{tabular}{c|rrrrrrrr|rrr}
            \toprule
                &\multicolumn{8}{c|}{FT-ScoreNets are trained on datasets \cite{nds, nb201, nb101, macro}}& \multicolumn{3}{c}{Handcrafted} \\
                \cmidrule(r){2-9} \cmidrule(r){10-12}
                & DARTS & NASNet & Amoeba & ENAS & PNAS & NB201 & NB101 & Macro & Zen-Score \cite{Zen-NAS}& Params & NASWOT \cite{naswot}\\
                \midrule
                DARTS & 0.774 & \textbf{0.675} & \underline{0.656} & 0.540 & \underline{0.659} & 0.422 & -0.308 & -0.149  &0.448 & 0.668 & 0.647\\
                NASNet & \textbf{0.633} & 0.742 & \textbf{0.584} & \textbf{0.644} & \textbf{0.557} & 0.329 & -0.011 & 0.240 &0.102 & 0.411 & 0.418\\
                Amoeba & \textbf{0.689} & \textbf{0.674} & 0.720 & \textbf{0.540} & \textbf{0.654} & 0.268 & -0.150 & 0.186 &-0.046 & 0.343 & 0.276\\
                ENAS & \textbf{0.701} & \textbf{0.637} & \textbf{0.579} & 0.676 & \textbf{0.594} & 0.472 & -0.062 & 0.096& 0.232 & 0.561 & 0.532\\
                PNAS & \textbf{0.666} & \textbf{0.567} & \textbf{0.626} & 0.431 & 0.658 & 0.237 & -0.199 & -0.126 &  0.259 & 0.541 & 0.496\\
                NB201 & 0.533 & -0.034 & -0.049 & -0.204 & 0.578 & 0.926 & -0.247 & 0.517 & 0.429 & 0.725 & 0.824\\
                NB101 & 0.288 & \underline{0.521} & 0.407 & \underline{0.473} & -0.250 & 0.082 & 0.854 & \underline{0.537}& 0.628 & 0.431 & 0.388\\
                Macro & \underline{0.761} & 0.570 & -0.276 & \underline{0.790} & \underline{0.787} & 0.612 & 0.600 & 0.979 &0.682 & 0.317 & 0.900 \\
                
            \bottomrule
          \end{tabular}
          \end{table*}
          
        \subsubsection{Score-Score Correlation}
            We measure the correlation between scores from different performance predictors. We report the correlation between our neural proxies and handcrafted proxies in Table \ref{tab:score_score_correlation} and Table \ref{tab:score_accuracy_correlation_imagenet}. Table \ref{tab:score_score_correlation} and Table \ref{tab:score_accuracy_correlation_imagenet} show score-score correlation of models, measured using Spearman's rank coefficient. Cell (row $i$, column $j$) indicates score-score correlation of model (neural \& handcrafted) $i$ and handcrafted model $j$ averaged on 8 datasets (CIFAR), 5 datasets (ImageNet).
            
            \begin{table}
                \caption{Score-score correlation of models on CIFAR}
                \label{tab:score_score_correlation}
                \centering
                \begin{tabular}{c|r|rrr}
                \toprule
                   & Models & Zen-Score & Params & NASWOT\\
                    \midrule
                    \multirow{8}{*}{FT-ScoreNets} 
                     & DARTS   & 0.064 & 0.301 & 0.312 \\
                     & NASNet  & 0.140 & 0.298 & 0.303 \\
                     & PNAS    & 0.115 & 0.333 & 0.318 \\
                     & ENAS    & 0.106 & 0.327 & 0.323 \\
                     & Amoeba  & 0.107 & 0.344 & 0.364 \\
                     & NB201   & -0.118 & -0.013 & 0.034 \\
                     & NB101   & 0.064 & 0.301 & 0.312 \\
                     & Macro   & -0.029 & 0.081 & 0.078 \\
                    \midrule
                    \multirow{3}{*}{Handcrafted}
                     & Zen-Score \cite{Zen-NAS} & 1.000 & 0.602 & 0.710\\
                     & Parameters & 0.602 & 1.000 & 0.803\\
                     & NASWOT \cite{naswot} & 0.710 & 0.803 & 1.000 \\
                \bottomrule
              \end{tabular}
            \end{table}

            \begin{table}
                \caption{Score-score correlation of models on ImageNet}
                \label{tab:score_accuracy_correlation_imagenet}
                \centering
                \begin{tabular}{c|r|rrr}
                \toprule
                    & Models & Zen-Score & Params & NASWOT\\
                    \midrule
                     \multirow{5}{*}{FT-ScoreNets} &DARTS & 0.084 & 0.510 & 0.438 \\
                    &NASNet & 0.171 & 0.389 & 0.413 \\
                    &PNAS & 0.045 & 0.426 & 0.349 \\
                    &ENAS & 0.330 & 0.557 & 0.606 \\
                    &Amoeba & -0.208 & 0.301 & 0.217 \\

                    \midrule
                    \multirow{3}{*}{Handcrafted}&Zen-Score \cite{Zen-NAS} & 1.000 & 0.500 & 0.676 \\
                    &Parameters & 0.500 & 1.000 & 0.742 \\
                    &NASWOT \cite{naswot} & 0.676 & 0.742 & 1.000 \\
                \bottomrule
              \end{tabular}
            \end{table}

        \subsubsection{End-to-End NAS} \label{sec:E2ENAS}
            We conducted End-to-End NAS Procedure on ResNet-like search space proposed by Zen-NAS and achieved competitive results against other zero-shot performance predictors in CIFAR-100. This search space consists of over $10^{99}$ architectures. We used our DE-NSGA2 search algorithm combined with the zero-shot performance predictors discovered by training on 8 search spaces and ensembling 8 neural performance predictors. We report the test-accuracy of the architectures found by our zero-shot performance predictors in comparison to architectures found by other zero-shot performance predictors in Table \ref{tab:E2E}. The results of SynFlow, TE-NAS \cite{TE-NAS}, NASWOT \cite{naswot}, Zen-NAS score are taken from Zen-NAS \cite{Zen-NAS}. Net$^2$ (1) is discovered using model trained via iteration method. Net$^2$ (2) is discovered using model trained via ensemble method. The parameter budget for models is 1 million parameters.

            We configure the NSGAII-DE algorithm to have both simple genetic algorithm operators for non-ordered variables and differential evolution operators for ordered variables. Crossover type \& probability and mutation rate for simple genetic operators are UX, $0.5$, and $0.8$, respectively. Crossover probability and differential weight for differential evolution operators are $0.8$, and $0.8$, respectively. The hyperparameters for simple GA operators are inspired from \cite{Genetic}. The maximum number of blocks and model size are 18 and 1 million, respectively. We consider block types (SuperResKXKX, SuperResK1KXK1), kernel size $(3, 5, 7)$, and stride $(1, 2)$ to be non-ordered. We consider the number of channels $(8, 16, \ldots, 2048)$, the number of bottleneck channels $(8, 16, \ldots, 256)$, and the number of sublayers in a block $(1, 2,\ldots, 9)$ to be ordered variables. The population size is 512 and number of generations is 100. We initialize the population to have the number of channels between 48 and 320, the number of bottleneck channels between 32 and 80, and the number of sublayers is 1 or 2. If the randomized model does not satisfy constraints on the maximum number of layers or model size, we ignore them. If the offspring does not satisfy the constraint, we multiply their fitness value with $-1$. Finally, we chose the model having at least 900,000 parameters (the budget is 1,000,000 parameters) with the highest score.

            \begin{table} 
                \caption{test-accuracy of architectures on CIFAR datasets}
                \label{tab:E2E}
                \centering
                \begin{tabular}{r|rr}
                \toprule
                    & CIFAR-10 & CIFAR-100\\
                    \midrule
                    Zen-NAS \cite{Zen-NAS} & 96.2\% & 80.1\% \\
                    SynFlow \cite{LightweightNAS} & 95.1\% & 75.9\% \\
                    TE-NAS \cite{TE-NAS} & 96.1\% & 77.2\% \\
                    NASWOT \cite{naswot} & 96.0\% & 77.5\%\\
                    Net$^2$ (1) & 96.6\% & 78.8\% \\
                    Net$^2$ (2) & 96.7\% & 77.5\% \\
                \bottomrule
            \end{tabular}
            \end{table}

        \begin{figure*}
            \centering
            
            \subfloat[Net$^2$ (1) architecture]{%
                \includegraphics[width=0.86\textwidth]{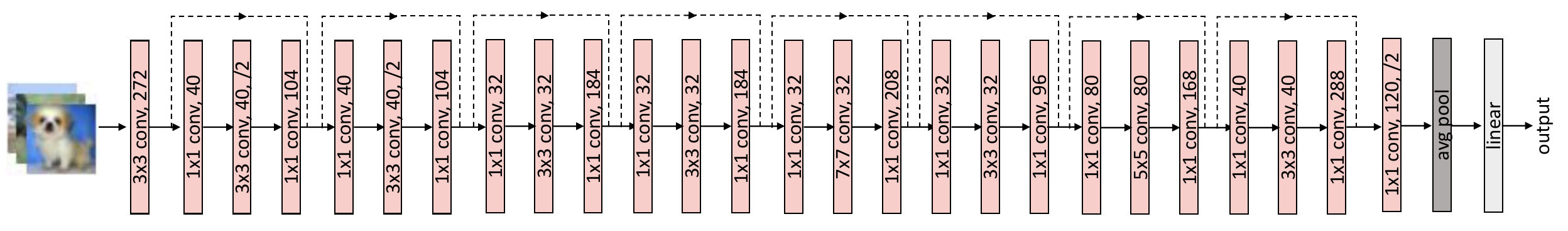}%
                \label{fig:search_full}%
            }
            
            
            \subfloat[Net$^2$ (2) architecture]{%
                \includegraphics[width=1\textwidth]{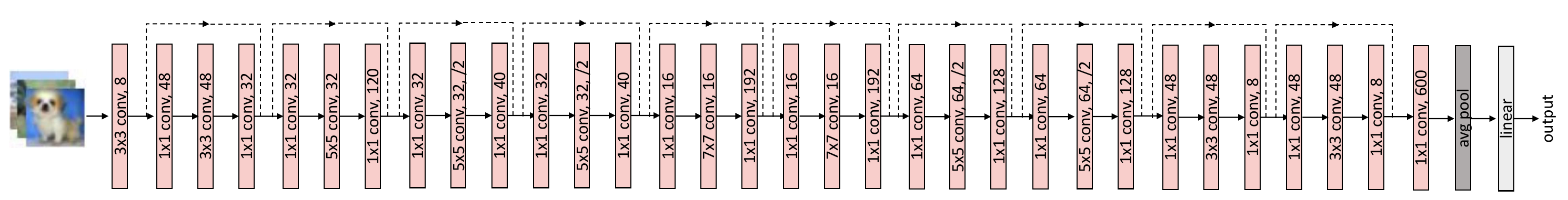}%
                \label{fig:search_ensemble}%
            }
            \vspace{1mm} 
            \caption{Architectures found by NSGAII-DE Algorithm and FT-ScoreNets}
            \label{fig:search}
        \end{figure*}

    \subsection{As a Prediction-based NAS}
        We conduct experiments on measuring score-accuracy correlation and End-to-End Prediction-based NAS procedure on NAS-Bench-201 \cite{nb201}. The model is trained on architecture-accuracy pairs using CIFAR-100 datasets and tested on architecture-accuracy pairs using CIFAR-10, CIFAR-100, and ImageNet16-120 datasets.
        \subsubsection{Score-Accuracy Correlation}
            In this experiment, we sampled 100 random architectures from NAS-Bench-201 as a dataset to train our model. Then, we use our model to measure the score-accuracy correlation of the scores from our model and test-accuracy of the remaining 15,525 architectures in NAS-Bench-201 (train: 100, test 15,525, total: 15,625). The NAS-Bench-201 score-accuracy correlation is depicted in Table \ref{tab:SAC_Prediction}. BRP-NAS, TNASP, and our method on prediction-based NAS on NAS-Bench-201 are featured in Table \ref{tab:SAC_Prediction}. BRP-NAS(1) predictor is transferred from the latency predicting task. BRP-NAS(2) predictor is not transferred from the latency predicting task. Our method uses an optimization process different from the previous two methods. The experimental results of BRP-NAS and TNASP are obtained from \cite{TNASP, BRP-NAS}. We trained the scorer with a sample size equal to the number of data points (100), using 200 training steps. The learning curve of the predictor is reported in Figure \ref{fig:learning_curve6}.

            \begin{table} 
                \caption{Prediction-based NAS on NAS-Bench-201}
                \label{tab:SAC_Prediction}
                \centering
                \begin{tabular}{r|rrr}
                \toprule
                    & Sample Size & Kendall-Tau & Spearman\\
                    \midrule
                    BRP-NAS(1) \cite{BRP-NAS} & 900 & - & 0.908 \\
                    BRP-NAS(1) \cite{BRP-NAS} & 100 & - & 0.890 \\
                    BRP-NAS(2) \cite{BRP-NAS} & 100 & - & 0.834 \\
                    TNASP \cite{TNASP} & 1563 & 0.726 & - \\
                    TNASP \cite{TNASP} & 78 & 0.565 & - \\
                    FT-ScoreNet & 100 & 0.749 & 0.904 \\
                \bottomrule
            \end{tabular}
            \end{table}
            
        \subsubsection{Searching on NAS-Bench-201}
            We search on NAS-Bench-201 using our trained model from 100 randomly sampled architectures. We compute scores of the remaining 15,525 architectures, then start training from the architecture with the highest score to the lowest. Within 10 trained architectures (combined with an additional 100 randomly sampled ones), we were able to reach $94.17\%$ on CIFAR-10, $73.01\%$ on CIFAR-100, and $46.06\%$ on ImageNet16-120.

\section{Discussion} \label{sec:discussion}
    \subsection{Score-Accuracy Correlation Task}
        From Table \ref{tab:score_accuracy_correlation}, we see that many neural-based models (with v-norm) outperform handcrafted scorers like Zen-Score, NASWOT-score, and number of parameters by a large margin. This indicates the ability of the model for cross-search-space generalizability. However, the models trained on NDS datasets do not outperform in the NAS-Bench-201, NAS-Bench-101, and NAS-Bench-Macro score-accuracy correlation compared to handcrafted scorers. We believe it is due to that:
            \begin{itemize}
                \item Final test-accuracy of models highly depends on their training settings.
                \item Architectures in NDS dataset across 5 search spaces use the same training settings.
                \item NDS dataset, NAS-Bench-201, NAS-Bench-101, and NAS-Bench-Macro uses different training settings from each other.
            \end{itemize}
        Training settings of datasets are described in Table \ref{tab:training_settings}. From Table \ref{tab:training_settings} and Table \ref{tab:score_accuracy_correlation}, we suspect the differences in the choice of optimizer impact the training result most.

        Table \ref{tab:score_accuracy_correlation} and table \ref{tab:score_accuracy_correlation_not_vnorm} show that the performance improvement when applying v-norm. The improvement is evidenced by that models without the v-norm algorithm exhibit 13 cases of negative correlation, whereas those with the algorithm only show 2 such cases. This showed that the information learned by models with v-norm algorithm is more likely to generalize to diverse search space, justifying the usage of the algorithm.

        Table \ref{tab:score_accuracy_correlation_imagenet} shows that the method can rank architectures for the ImageNet-Image-Classification task. NDS-ImageNet's dataset size is smaller than its NDS-CIFAR counterpart, 124 versus 5,000. This shows that the method is data-efficient by the result it achieves from learning a small dataset.
        
        \begin{table}
                \caption{Training settings for architectures in benchmarks}
                \centering
                \begin{tabular}{r|rrrr}
                \toprule
                    Training Setting & NB201 & NB101 & NDS\\ 
                    \midrule
                        Batch Size & 256 & 256 & 128 \\ 
                        Number of Epochs & 200 & 108 & 100 \\ 
                        Optimizer & SGD & RMSProp & SGD 
                        \\ 
                    \bottomrule
                    \end{tabular}
                \label{tab:training_settings}
            \end{table}

    \subsection{End-to-End NAS Task}
        In prediction-based NAS settings, Table \ref{tab:SAC_Prediction} shows that our method reaches 0.904 Spearman Correlation without the need of being transferred from latency predicting task while using merely multi-layer-perceptrons. This performance surpasses both the latency-transfered BRP-NAS \cite{BRP-NAS} and transformer-powered TNASP \cite{TNASP}. The NAS-Bench-201 End-to-End NAS procedure via brute-fore search/greedy testing strategy also shows promising results. While this does not promise a superior performance on prediction-based NAS settings on real architecture search space, this shows our encoding mechanism via Fourier Transform and our optimization method can efficiently exploit the data. We believe the efficiency of our representation mechanism comes from more information on architecture structure stored by our encoding than by one-hot encoding.

        In Zero-Shot NAS settings, Table \ref{tab:E2E} shows that our method discovered architectures under one million parameters having 96.7\% on CIFAR-10 and 78.8\% on CIFAR-100. This is a strong evidence that neural-based Zero-Shot NAS can perform well on real NAS search space. Within the same amount of parameters, Net$^2$ has 0.4\% - 0.5\% test-accuracy improvement on CIFAR-10 compared to best handcrafted method. However, the architectures did not outperform Zen-NAS in CIFAR-100. This is understandable as NDS, Macro, and NASBench101 used CIFAR-10 as the dataset for image classification task. We only use CIFAR-100 for NAS-Bench-201's score-accuracy correlation pairs. In the ensemble method, the weight for NAS-Bench-201 is less than $10^{-4}$. In the iteration method, each dataset is treated equally (which is the ratio of 7:1 for CIFAR-10:CIFAR-100). Hence, the learned model failed to craft architectures that can classify inner categories and often put an awkward small number of channels (8, see Figure \ref{fig:search_ensemble}) toward the last layers of architectures.
        
    \subsection{Other Insights}
        We also show the correlation between our neural scorers and handcrafted scorers for drawing insights on what our models learn and did not learn in Table \ref{tab:score_score_correlation} and Table \ref{tab:score_accuracy_correlation_imagenet}. In Table \ref{tab:score_accuracy_correlation}, a weak correlation between our FT-ScoreNets and handcrafted scorers is observed, in contrast to the moderate-to-strong correlation between handcrafted ones. In Table \ref{tab:score_accuracy_correlation_imagenet}, we see a low-to-moderate correlation between our FT-ScoreNets and handcrafted scorers. This indicates a trend where with more data points, FT-ScoreNets can see distinct useful patterns that have not been fully understood before. Understanding FT-ScoreNets' pattern is an open question for future work.

        Figure \ref{fig:learning_curve} shows that in Zero-Shot NAS data-scarce settings, training longer does not always result in better performance. However, in Prediction-based NAS settings, training longer results in higher correlation. This might indicate that the neural scorers overfit to the structure of the search space instead of learning meaningful features. This overfitting phenomenon also explains why scorers training on NAS-Bench-201 and NAS-Bench-101 perform so badly compared to other scorers (Table \ref{tab:score_accuracy_correlation}). While NAS-Bench-201 and NAS-Bench-101 datasets are large (about $15K$ and $500K$ architectures), the architectures within them lack of diversity.
    
            
        
\section{Conclusion} \label{sec:conclusion}
    In this paper, we showed an effective application of Neural Prediction-based NAS in Zero-Shot NAS scenarios. We proposed a novel representation mechanism, employing a DFT-based method specifically tailored for convolutional operations. In addition, we presented an innovative optimization scheme, flexible for use in both singular and multiple search spaces. Our methodology not only led to the discovery of competitive architectures in large NAS-search spaces but also proved effective when evaluated as a prediction-based NAS. Through this study, we found that mechanisms like DFT which is not directly linked to neural network's strength or expressiveness, such as the Fourier Transform, can successfully differentiate architectures and aid neural networks in ranking them. This indicates the potential for future Zero-Shot NAS research in exploring novel encoding methods to differentiate neural networks. Future directions of deep Zero-Shot NAS might be the adaptation of deep learning techniques, solutions for domain shift in search space, new encoding techniques, or new NAS datasets for diverse tasks and types of architectures.
    


 






%

\appendices




\ifCLASSOPTIONcaptionsoff
  \newpage
\fi
\bibliography{reference}
\bibliographystyle{IEEEtran}

\end{document}